\documentclass[authoryear,review,12pt]{elsarticle}
\usepackage{courier}

\makeatletter
\def\ps@pprintTitle{%
 \let\@oddhead\@empty
 \let\@evenhead\@empty
 \def\@oddfoot{\textit{Accepted for publication in Expert Systems With Applications}\hfill}%
 \let\@evenfoot\@oddfoot}
\makeatother

\usepackage{amssymb}
\usepackage{url,multirow,morefloats,floatflt,cancel,tfrupee}
\usepackage{adjustbox}
\usepackage{booktabs}
\usepackage{lineno}
\usepackage[dvipsnames]{xcolor}
\definecolor{red}{rgb}{0,0,0}
\definecolor{brown}{rgb}{0,0,0}
\definecolor{blue}{rgb}{0,0,0}

\journal{Expert Systems with Applications}

\begin{document}

\begin{frontmatter}



\title{Deep 1D-Convnet for accurate Parkinson disease detection \textcolor{brown}{and severity prediction} from gait }


\author[poly]{Imanne El Maachi}
\author[poly]{Guillaume-Alexandre Bilodeau}
\author[teluq]{Wassim Bouachir}

\address [poly]{Polytechnique Montreal,\\ 2900 boul. Edouard Montpetit, Montreal (Qc), H3T 1J4, Canada \\ \texttt{\{imanne.el-maachi, guillaume-alexandre.bilodeau\}@polymtl.ca} }

\address[teluq]{T\'{E}LUQ University,\\ 5800 rue Saint-Denis, bur. 1105, Montreal (Qc), H2S 3L5, Canada\\ 
\texttt{wassim.bouachir@teluq.ca}
}

\begin{abstract}

Diagnosing Parkinson's disease is a complex task that requires the evaluation of several motor and non-motor symptoms. During diagnosis, gait abnormalities are among the important symptoms that physicians should consider. However, gait evaluation is challenging and relies on the expertise and subjectivity of clinicians. In this context, the use of an intelligent gait analysis algorithm may assist physicians in order to facilitate the diagnosis process.
This paper proposes a novel intelligent Parkinson detection system based on deep learning techniques to analyze gait information.
We used 1D convolutional neural network (1D-Convnet) to build a Deep Neural Network (DNN) classifier. The proposed model processes 18 1D-signals coming from foot sensors measuring the vertical ground reaction force (VGRF). The first part of the network consists of 18 parallel 1D-Convnet corresponding to system inputs. The second part is a fully connected network that connects the concatenated outputs of the 1D-Convnets to obtain a final classification. 
We tested our algorithm in Parkinson's detection and in the prediction of the severity of the disease with the Unified Parkinson's Disease Rating Scale (UPDRS).
Our experiments demonstrate the high efficiency of the proposed method in the detection of Parkinson disease based on gait data. The proposed algorithm achieved an accuracy of $98.7\%$. To our knowledge, this is the state-of-the-start performance in Parkinson's gait recognition. Furthermore, we achieved an accuracy of $85.3 \%$ in Parkinson's severity prediction. To the best of our knowledge, this is the first algorithm to perform a severity prediction based on the UPDRS.

These results show that the model is able to learn intrinsic characteristics from gait data and to generalize to unseen subjects, which could be helpful in a clinical diagnosis.

\end{abstract}

\begin{keyword}
1D-Convnet \sep Parkinson \sep gait \sep  Classification \sep Deep learning



\end{keyword}

\end{frontmatter}

\section{Introduction}

Today, over 10 millions people suffer from Parkinson's disease. Unfortunately, no cure exists to heal this disorder. That is why early diagnosis is important to improve the patient's treatment. Currently, physicians evaluate symptoms such as shaking, difficulty to initiate movements, slowness, and difficulty to walk \textcolor{blue}{\citep{jankovic2008parkinson}}. \textcolor{red}{One of the most used tools in Parkinson clinical evaluation is the Unified Parkinson's Disease Rating Scale (UPDRS). This scale consists of 42 criteria/questions that cover different aspects of Parkinson's disease. These aspects comprise motor symptoms (including gait features), behavioral characteristics, and daily activities \citep{fahn1987updrs}. For the most part, doctors evaluate the severity of the criteria on a scale of 0 (normal) to 5 (severe). The total score is the sum over all the criteria (maximum is 176).}

\textcolor{blue}{Gait analysis is an important step in the diagnosis process of Parkinson's disease \citep{jankovic2008parkinson}\citep{REICH2019337}, as gait abnormalities have been documented to occur at the early stages \citep{pistacchi2017gait}}. The Parkinsonian gait is mainly characterized by small steps, a slower gait cycle, \textcolor{blue}{an increase in stride variability}, a shorter swing phase, a longer stance phase and a flat foot strike instead of toe-to-heel strike \citep{morris2001biomechanics} \citep{perumal2016gait}. \textcolor{blue}{Physicians evaluate these features in their diagnosis process to confirm the presence of the Parkinson disease \citep{jankovic2008parkinson}. However, gait's evaluation can be challenging since it can be affected by several factors such as age and health condition.} 

Despite the significant interest in parkinsonian gait analysis, there is no objective tool to assist physicians for gait evaluation. Since changes in the gait are among the first symptoms of this disease \citep{pistacchi2017gait}, a powerful gait classifier would be helpful for physicians. \textcolor{blue}{In this clinical context, the objective of our research work is to develop an intelligent tool to detect Parkinson's disease symptoms \textcolor{red}{and predict the Parkinson severity rate (based on the UPDRS)} from gait data.}



To detect these characteristics, feature extraction methods have been widely used \citep{ertuugrul2016detection, daliri2013chi, sarbaz2012gait, xia2015classification, mannini2016machine, wang2017machine}. Previous studies used temporal \citep{ertuugrul2016detection, xia2015classification, wang2017machine, mannini2016machine} or frequential \citep{daliri2013chi, sarbaz2012gait} tools to get differentiable patterns between normal and parkinsonian gait. However, gait is a physiological characteristic that differs for each person according to age, health, and other intrinsic factors. Therefore, manual preprocessing and feature extractions will always be limited in their capacity. To avoid hand-crafted signal processing, we propose a novel gait classifier based on deep learning without explicit feature extraction. The proposed Deep Neural Network (DNN) is composed of two parts. The first one consists of 18 parallel 1D-Convnets, each processing a vertical ground reaction force (VGRF) signal coming from a foot sensor. Each 1D-Convnet extracts deep features that are subsequently concatenated together. The second part is a fully connected network that processes the concatenated vector to output the final decision. Thanks to its ability to extract relevant gait features from different input signals, the proposed model outperforms state-of-the-art methods by achieving an accuracy of 98.7\%. 
 
The main contribution of this work is \textcolor{red}{threefold}. Firstly, we developed a 1D-Convnets that extracts relevant deep features for accurate gait classification, which avoids manual feature extraction. Secondly, we show that our proposed method \textcolor{red}{achieves state-of-the-art accuracy (98.7 \%) in Parkinson's detection. Thirdly, we present the first algorithm in UPDRS severity prediction, which is extremely valuable for clinical decision support systems, and achieves an accuracy of 85.3 \%. The source code of this project is publicly available to ensure reproducibility for future research \footnote{The source code is available at https://github.com/imanneelmaachi/Parkinson-disease-detection-and-severity-prediction-from-gait}.}

The rest of the paper is organized as follows. In section \ref{sec:litt}, we present a review of previous works on gait classification, as well as background concepts. Section \ref{sec:meth} introduces the proposed method. Section \ref{sec:exp} describes the experimental setup. Section \ref{sec:res} presents results and discussion, and we finally conclude in section \ref{sec:conc}.

\section{Related work and background}\label{sec:litt}

\subsection{Previous gait classification methods}
Various feature extraction and classification approaches have been explored in previous work for gait analysis. Some of them extracted temporal patterns, while others used frequential features. 
In the temporal domain, \citet{ertuugrul2016detection} proposed an algorithm based on shifted 1D local binary patterns (1D-LBP) with machine learning classifiers. They used 18 VGRF input signals coming from foot sensors of Parkinson patients and control subjects. For each signal, they applied shifted 1D-LBP to construct 18 histograms of the 1D-LBP patterns from which they extracted statistical features, such as entropy, energy and correlation. Finally, they concatenated the features from all the 18 histograms and used various supervised classifiers, such as Random Forest and Multi-Layer Perceptron (MLP), to classify the feature vectors. In the frequential domain, \citet{daliri2013chi} applied a Short-Time Fourier transform (STFT) on each input signal using the same VGRF input data. Then, they extracted the mean frequency and the variance of the frequency to construct a histogram representing the distribution of features extracted from all input signals. This histogram was processed with a feature discriminant ratio (FDR) in order to select the most significant bins. The final bins were then classified by a Support Vector Machine (SVM) with a chi-square kernel. 

For other gait disorders, \citet{mannini2016machine} first trained a Hidden Markov Model (HMM) on acceleration signals recorded with sensors from elderly, post-stroke and Huntington's disease patients. Then, they constructed a feature vector using the HMM log-likelihood of each patient with each class, combined with various temporal and frequency features. The final vector was sent to an SVM classifier. \citet{wang2017machine} proposed a similar algorithm, but instead of using an HMM model to extract features, they trained a K-Nearest Neighbor (KNN) classifier and then used the probability given by that classifier for each class as a feature. An SVM was later trained to classify feature vectors containing those probabilities with other statistics. 
\citet{xia2015classification} used spatiotemporal features of neurodegenerative disease patients and control subjects. For each time series, they computed various statistics, such as the fuzzy entropy, skewness, and kurtosis. They used machine learning classifiers, such as Random Forest, SVM, MLP, and KNN. After an optimization with a features selection algorithm, the best result was obtained with an SVM. 

Recently, \citet{zhao2018hybrid} developed a deep learning algorithm to detect Parkinson disease. Their model was composed of two parallel networks. The first network analyzed the spatial distribution of forces with a 2D-Convnet. The second network analyzed the temporal distribution with a recurrent neural network (RNN). The final classification was decided by the average of both channels. 

\subsection{Background concepts}
The method that we propose is based on deep learning. To be self-explanatory, some basic concepts are briefly explained in this section. 
The learning in DNN relies on intermediate layers, commonly called hidden layers. The input is processed at each layer in order to reach higher abstraction levels. During the training stage, the DNN increases its accuracy by using stochastics optimizers that decrease the loss at each iteration. Then, with enough data, the DNN learns how to extract features from the input \citep{goodfellow2016deep}.
Each layer is composed of neurons. The neuron computes a weighted summation of the inputs to which it is connected. A nonlinear activation function is then applied to obtain the neuron output. The learning is based on the adaptation of the weights W: The deep network learns the optimal W values to obtain the closest estimate to the real value \citep{goodfellow2016deep}. In order to avoid over-fitting, regularization techniques, such as dropout, are commonly used. Dropout reduces the over-fitting by shutting off some neurons chosen randomly at each iteration during the training \citep{srivastava2014dropout}.
In our model, we use convolutional layers, max-pooling layers, and fully-connected layers. Each type of layer is briefly described:

\begin{itemize}
    \item \textbf{Convolutional layers}: they perform a spatial convolution between the inputs and the filters. The filters contain weights which are the learned elements of the layer. 
    \item \textbf{Max-pooling layers}: they sample the input to form an output with smaller dimensions by selecting the maximum value element.
    \item \textbf{Fully connected layers (FC)}: they connect the output of the convolution layers to the final output. They allow a convergence of the decision at a higher level.
\end{itemize}

\section{Methods} \label{sec:meth}

\subsection{Method overview}
\begin{figure*}[!t]
    \centering
    \includegraphics[trim=160 180 50 180,clip, scale = 0.5]{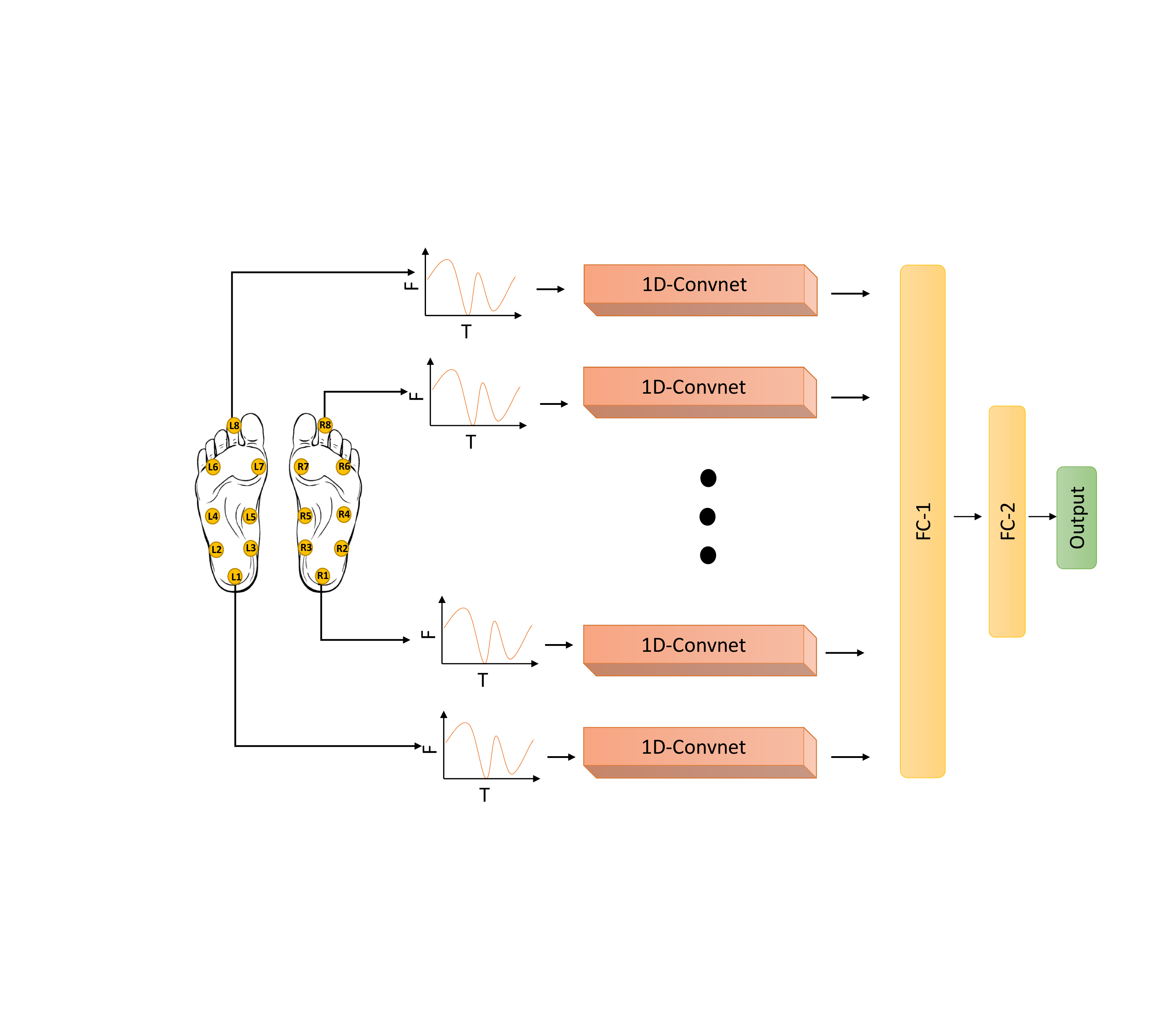}
    \caption{The architecture of the DNN: The first part is composed of 18 parallel 1D-Convnets, each processing a VGRF signal (F) coming from foot sensors. The 1D-Convnets are followed by fully connected layers that produce the final classification}
    \label{fig:shema}
\end{figure*}

The objective of our algorithm is to classify each subject's walk into one of the two categories: Parkinson and control. Each recorded walk was documented with 18 VGRF 1D signals, measured in Newtons as a function of time, from several foot sensors. \textcolor{red} { These signals correspond to 8 sensors placed underneath each foot (figure \ref{fig:shema}), in addition to two more signals that represent the sum of the 8 VGRFs for each foot.} \textcolor{red} {The VGRF signals comprise relevant information for gait analysis and characterization. In fact, important clinical spatio-temporal gait features, such as swing phase, stance phase and stride time, can be derived from VGRF signals. Feeding our deep learning model with raw data representing vertical ground reaction force records will allow to go beyond the hand-crafted features, by implicitly extracting relevant features for the studied application.} 

We divided VGRF signals into $m$ segments labeled with the subject category. These segments are the input samples of our DNN. The DNN is composed of two parts (figure \ref{fig:shema}). The first part consists of 18 parallel 1D-Convnets, while the second part is a fully connected network that operates on the concatenation of the features extracted by the 18 1D-CNN. The final walk classification was decided according to the majority classification of all the subject segments. 

\subsection{Parallel 1D-Convnet branches}
\begin{figure*}
    \centering
    \includegraphics[trim=10 125 10 100,clip,width=\linewidth,scale=0.7]{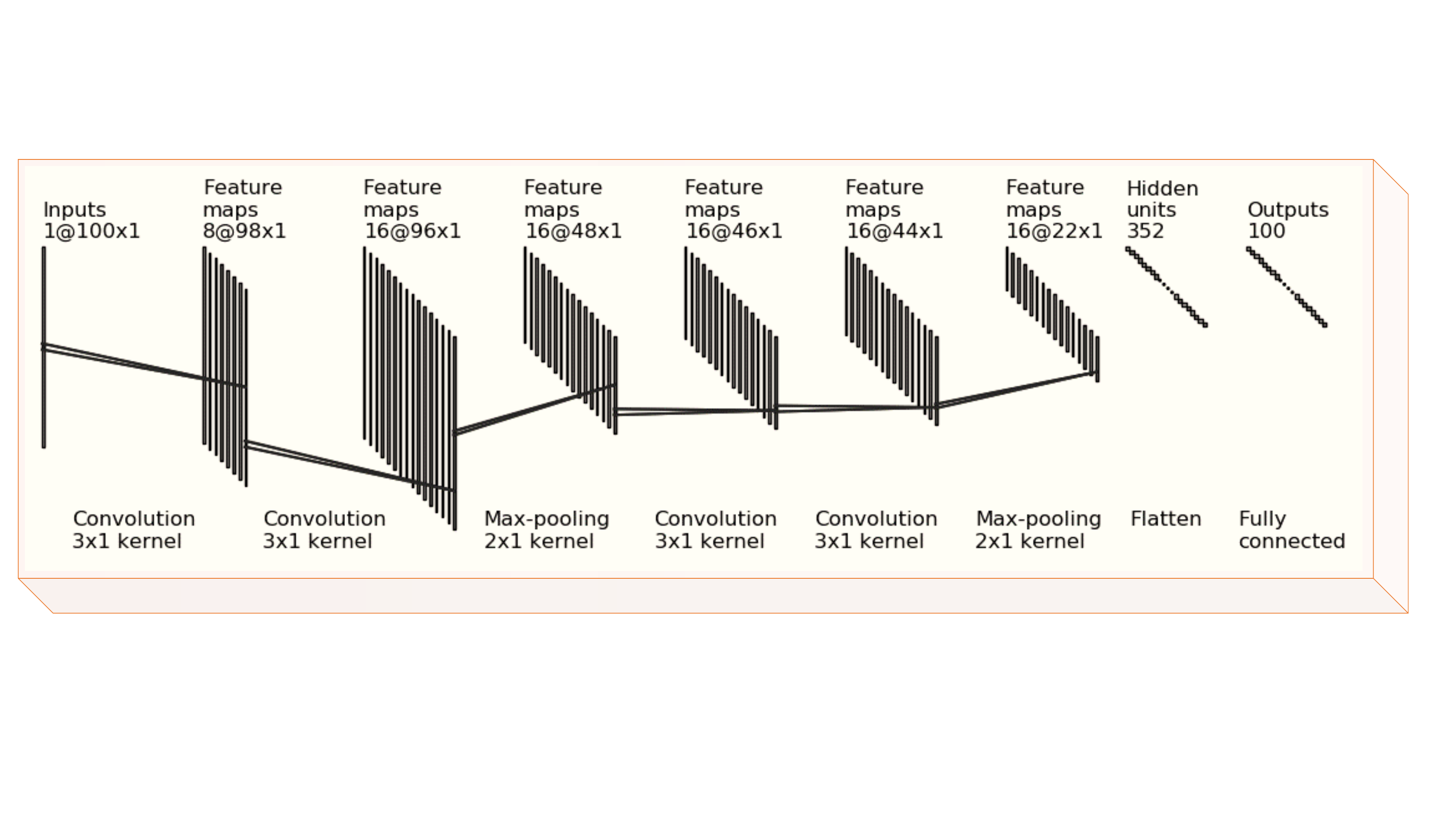}
    \caption{1D-Convnet: Each 1D-Convnet block is composed of 4 convolutional layers, 2 max-pooling layers and 1 fully-connected layer. }
    \label{fig:cnn}
\end{figure*}

The first part of the network is composed of 18 parallel 1D-Convnets (figure \ref{fig:cnn}). Each network takes as input a VGRF signal going through 4 convolutional layers, that lead to a fully connected layer. Every two convolutional layers is followed by a max-pooling layer. Each filter in the convolutional layers extracts a specific pattern that allows differentiating between the Parkinson and control group. This 1D-Convnets parallelization allows the treatment of each signal independently. In fact, since each sensor records a specific data from a specific point, each time series has its own deep features and is analyzed separately with the 1D-Convnets parallelization.

\subsection{Fully connected bloc}
The second part of the network is designed to learn the relationship between the spatial feature extracted from the 1D-Convnets and the final output classification. First, the outputs of those 18 parallel 1D-Convnets are all concatenated into one single vector. This deep feature vector passes through two fully connected layers leading to the output layer. \textcolor{red}{For the Parkinson's detection,} the output layer is composed of one neuron that predicts the classification probability. \textcolor{red}{For the severity prediction, the output layer is composed of 5 neurons (for 5 classes) to predict the final classification.} This design has two main advantages: \begin{itemize}
    \item Selection: the network selects the most significant time series that contribute the most to the performance of the algorithm.
    \item Abstraction: time series from different sensors are \textcolor{blue}{merged together} only after passing through the 1D-Convnets. This allows getting significant features at the concatenation step. \textcolor{blue}{In fact, at this hidden state, each feature vector is obtained after multiple convolution steps. Therefore it would contain more abstract/global features. In our case, we believe that merging global features together is more efficient than merging raw data at the input level, since it is more robust to noise and local changes.}
\end{itemize}

\section{Experiments}\label{sec:exp}

\subsection{Database}
We used the public database collected by \textcolor{blue}{Physionet} \footnote{ https://physionet.org/content/gaitpdb/1.0.0/}, which includes data reported by \citet{frenkel2005effect, frenkel2005treadmill}, \citet{yogev2005dual} and \citet{hausdorff2007rhythmic}. In total, 93 parkinsonian patients and 73 control subjects participated in those studies. \textcolor{blue}{Each subject walked on a flat floor for two minutes at their natural pace. For each subject, eight sensors were placed under each foot (figure \ref{fig:shema}). Each sensor measures the VGRF (in Newton)
under its specific point. The sampling frequency was 100 samples per second.} The dataset also includes double-task time series, where subjects were walking while doing another activity. This results in an \textcolor{brown}{unbalanced 305} walks recorded  (70\% Parkinson and 30\% control walks). For each walk, 18 time series signals are available: 16 ($8 \times 2$) VGRF recorded from 8 sensors on each foot and 2 total VGRFs under each foot.
\textcolor{red}{For each subject (Parkinson or control), the UPDRS total score is reported in the database. The scores are distributed in a range between 0 and 70. }


\subsection{Parkinson's severity prediction}
\textcolor{red} {We have also tested our algorithm to predict the severity of the Parkinson's disease. We used the Unified Parkinson's Disease Rating Scale (UPDRS), which is the most used scale for Parkinson severity rating scale \citep{ramaker2002systematic}. We segmented the continous scale into 5 levels : \begin{itemize}
    \item class 1: UPDRS $<$ 5
    \item class 2: 5 $\leq$ UPDRS $<$ $15$
    \item class 3: 15 $\leq$ UPDRS $<$ 25 
    \item class 4: 25 $\leq$ UPDRS $<$ 35 
    \item class 5: 35 $\leq$ UPDRS
\end{itemize}
We kept the same architecture of the DNN and the same training hyperparameters. We only modified the last layer. We replaced it with a fully connected of 5 neurons, with a softmax activation layer.}
\subsection{Evaluation metrics}
\textcolor{red}{We used cross-validation with 10 folds to test our algorithm on 300 walks. We divided each of the Parkinson and the control group into 10 folds at the subject-level. This allowed us to keep the same dataset balance (70 \% Parkinson - 30\% Control) for each fold. We report the evaluation metrics over the combined validation predictions.}
\subsubsection{Parkinson and control subject classification}
The control group is identified as the negative (N) group and the Parkinson group is the positive (P) group. We use the following notations:
\begin{itemize}
    \item Number of true positive (TP): Parkinson subject correctly classified 
    \item Number of true negative (TN): control subject correctly classified 
    \item Number of false positive (FP): control subject misclassified 
    \item Number of false negative (FN): Parkinson subject misclassified 
\end{itemize}

The specificity (Sp), sensitivity (Se), and accuracy (Acc) are calculated as follows: 
\begin{equation}
Se = \frac{TP }{TP +FN} \end{equation}
\begin{equation}
Sp = \frac{TN }{TN +FP} \end{equation}

\begin{equation}
Acc = \frac{TP + TN }{TP+TN+ FP+FN}
\end{equation}

\subsubsection{Parkinson severity prediction}

Since this is a multiclassification experiment, we report the precision, the recall, and F1 score for each class. For each class, true positives (TP) are the patients with the corresponding label who are correctly classified. 
\begin{equation}
Precision = \frac{TP }{TP + FP} \end{equation}
\begin{equation}
Recall = \frac{TP }{TP +FN} \end{equation}

\begin{equation}
F1 = 2 \times \frac{Precision \times Recall }{Precision + Recall}
\end{equation}

\subsection{Training}
Since deep learning algorithms require large datasets, each walk was divided into smaller segments of 100 time steps with 50\% overlap. Each segment was labeled with the subject category for the training. \textcolor{red}{This segmentation was done inside each fold. Thus, the segments of a given subject are never divided between the training and validation set. The whole dataset contained 64468 segments. } %

The deep network was trained to classify these segments. \textcolor{red}{For Parkinson's disease detection,} the output probability of each segment was binarized (control or Parkinson). Then, the preponderant class ($>$50\%) for the given segmented whole walk determined the final \textcolor{red}{subject} classification. \textcolor{red}{For the severity prediction, the mode class over all the segments defined the subject's UPDRS class.}

 The final hyperparameters of the DNN are presented in table \ref{table:hyperp}. \textcolor{red}{A Scaled Exponential Linear Unit (SeLu)} activation function \citep{klambauer2017self} was used for all the hidden layers and a sigmoid activation \textcolor{blue}{(threshold = 0.5)} was used for the final output. The algorithm was trained using \textcolor{red}{Nesterov Adam optimizer \citep{dozat2016incorporating}} with a batch size of 800 and an initial learning rate of 0.001. We used early stopping to avoid over-fitting: if the validation accuracy did not improve for 10 epochs, we stopped the training and went back to the weights that gave the best validation accuracy. From there, we decreased the learning rate by a factor of 2 and continued the training. This process was repeated \textcolor{red}{4} times to complete the training of the model. 
 Also, in order to avoid over-fitting, some dropout was applied to the DNN to regularize the training.

\begin{table*}[!t]
\centering 
\caption{Layer descriptions \textcolor{red}{for Pakinson's disease detection}: the first part of the DNN is composed of 18 1D-Convnet, and the second part is a fully connected network.}
\begin{adjustbox}{width=\columnwidth,center}
\begin{tabular}{|l|l|l|l|l|l|l|}
\hline
                  & Layer no & Layer type & Number of units & Kernel size & Dropout \\ \hline
\multirow{8}{*}{1D-Convnet $\times 18$ } & 1  & Convolutional & 8    & 3 & 0 \\ \cline{2-6} 
                  & 2  & Convolutional  & 16 & 3    & 0       \\ \cline{2-6} 
                  & 3  & Max-pooling & - & 2    & 0       \\ \cline{2-6}
                  & 4  & Convolutional & 16    & 3 & 0 \\ \cline{2-6} 
                  & 5  & Convolutional  & 16 & 3    & 0       \\ \cline{2-6} 
                  & 6  & Max-pooling & - & 2    & 0       \\ \cline{2-6} 
                  &  7 & Flatten  & - & -    & -       \\
                  \cline{2-6} 
                  & 8  & FC  & 100 & -    & 0.5       \\
                  \hline
                  
\multirow{4}{*}{FC}   & 9  & Concatenate & -   & - & 0.5 \\ \cline{2-6}
                &  10 & FC & 100   & - & 0.5 \\ \cline{2-6} 

                 & 11 & FC & 20  & - & 0.5    \\ \cline{2-6} 
                  & 12 &  Output & 1  & - & -  \\
\hline 
\end{tabular}
\end{adjustbox}
\label{table:hyperp}
\end{table*}

\subsection{Models used for comparison}
\textcolor{red}{For Parkinson's detection,} we compared our method with the work of \citet{ertuugrul2016detection} and \citet{zhao2018hybrid}. \citet{ertuugrul2016detection} work was chosen for comparison because they used a simple manual feature extraction method that achieved accuracy results comparable with other machine learning algorithms. Since \citet{ertuugrul2016detection} used the same evaluation setting (10-folds cross-validation), we reported results from their paper. They had employed a feature extraction algorithm summarized as follows: 

\begin{itemize}
    \item For each signal (18 in total): \begin{itemize}
        \item Extract 1D-LBP with 8 neighbors. The output has the same length as the input with values between 2\textsuperscript{0} and 2\textsuperscript{8}. 
        \item Construct a histogram of the frequency of the obtained LBP. 
        \item Extract statistical features from the histogram such as skewness, entropy, energy, a coefficient of variation, kurtosis and correlation. 
    \end{itemize}
    \item Concatenate the features of all the histograms. 
    \item Train a machine learning classifier with the feature vectors created. 
\end{itemize}

We report their results for three classifiers: Naive Bayes (NB), Random Forest (RF) and the Multi Layer Perceptron (MLP). 

\citet{zhao2018hybrid} work was chosen because they also use deep learning algorithms, but with a different network architecture. 
\textcolor{brown}{We reproduced \citet{zhao2018hybrid} algorithm (according to the authors' paper description) in order to compare the algorithm in the same evaluation set-up with 10-fold cross-validation}. As described before, they have employed a two-channel network: a 2D-Convnet and an RNN. The 2D-Convnet was built with 2 convolutional layers, 2 max-pooling layers, an FC layer, and an output classification layer. The RNN was built with two Long short-term memory (LSTM) layers, an FC layer, and an output classification layer. The final classification was made by the average of both channels. More implementation details are available in their paper \citep{zhao2018hybrid}. They used the same 18 VGRF features, to which they added a time vector. Each input sample was an array of 19 features $\times$ 100 time steps. Their final output is a classification of the sample input. In order to evaluate their algorithm on a subject base, we took the majority classification of all the segmented walk samples to determine the subject final classification (as we do with our algorithm). 

We used Keras with the Tensorflow backend and the sklearn library for model implementation.

\section{Results \& Discussion}\label{sec:res}

\subsection{Evaluation of the algorithm}

\begin{figure*}
    \centering
    \includegraphics[trim=10 40 10 0,clip,width=\linewidth,scale=1.3]{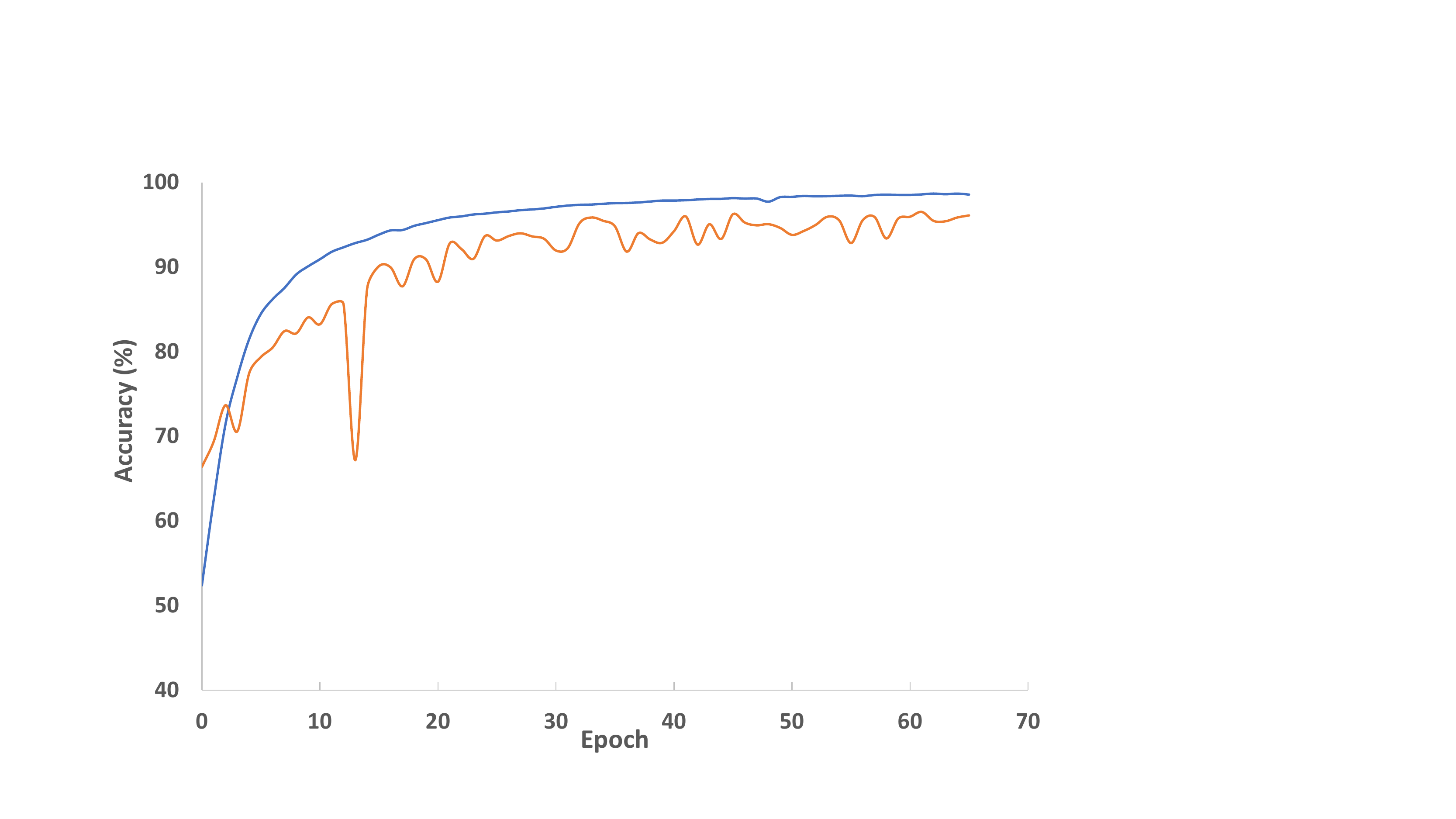}
    \caption{Training curve \textcolor{red}{for Parkison's detection}. The accuracy of the training set is represented by the blue line and the accuracy of the validation set is represented by the orange line. The accuracy corresponds to the proportion of segments correctly classified over all the walking segments in the training or the validation set.}
    \label{fig:training}
\end{figure*}

\begin{table*}[t]

\centering
\caption{Cross-validation results \textcolor{red}{for Parkison's detection}. Best results are in bold and the second best results are in italic. \textcolor{brown}{Sp: Specificity, Se: Sensibility, Acc: Accuracy, SD: Standard deviation.} Not available values for results reported in \citep{ertuugrul2016detection} are indicated by n/a. }
\begin{adjustbox}{width=\columnwidth,center}
\begin{tabular}{|l||c|c|c|c|c|c|c|c|c|c|c|}
\hline
 Algorithm  &  TP & FN & TN & FP & Sp $\pm$ SD (\%)  & Se $\pm$ SD (\%) & Acc $\pm$ SD(\%)\\
\hline\hline
Proposed DNN  & 206 & 4 & 90 & 0 &\textbf{100.0 $\pm$ 0.0} & \textbf{98.1 $\pm$ 3.3 } & \textbf{98.7 $\pm$ 2.3} \\\hline
DNN \citep{zhao2018hybrid} (Reproduction) & 202 & 8 & 69 & 21 & 76.7 $\pm$ 8.2  & \textit{96.2 $\pm$ 3.8} & \textit{90.3 $\pm$ 2.9} \\\hline

MLP \citep{ertuugrul2016detection}& n/a & n/a & n/a & n/a & \textit{82.2} & 88.9 & 88.9  \\ \hline

NB \citep{ertuugrul2016detection}& n/a & n/a & n/a & n/a & n/a & n/a& 76.1  \\ \hline

RF \citep{ertuugrul2016detection}& n/a & n/a & n/a & n/a & n/a & n/a& 86.9  \\ \hline

\end{tabular}
\end{adjustbox}
\label{table:test}
\end{table*}


\textcolor{red}{The training curve illustrated in figure \ref{fig:training} shows that our algorithm can converge quickly. Furthermore, the accuracy of the validation set increases as the training accuracy increases, which shows that the algorithm is learning meaningful features without over-fitting.}
 The performance of our method is compared with other studies in table \ref{table:test}. \textcolor{blue}{ Recall that for deep learning algorithms, the result for a subject is obtained by a majority vote over the classification of gait segments. At the segment level, we have an accuracy of 98.3\%, a specificity of 99.2\% and a sensitivity of 97.8\%.} \textcolor{red}{Our proposed DNN showed an accuracy of 98.7\% at the subject-level, which clearly outperforms previous algorithms.
 Compared to other methods, our algorithm has the advantage of processing multiple input signals differently. This allows it to extract the most meaningful and specific features from each signal.} 

\textcolor{red}{However, we notice a discrepancy between the results obtained for \citet{zhao2018hybrid} algorithm and the results reported in their paper (accuracy of 98,6\%). This could be explained by the different evaluation method (cross-validation vs 1 test fold).} \textcolor{red}{Our algorithm is different from \citet{zhao2018hybrid} by the independence between input signals. In our method, the independence between the input 1D signals allows the method to be generalized to other experimental settings. Our method can be easily adapted to more or fewer input signals. Thus, in a clinical context, our algorithm can be easily adapted to different gait clinical studies. In our research, we used this property of our algorithm to perform an ablation study between the input signals.}

Compared to classical machine learning models, our algorithm is more adapted to the gait classification problem. First, gait time series are non-linear and noisy signals. \textcolor{blue}{Our model is adapted to this type of data since it uses a deep learning approach. In fact, compared to classic machine learning algorithms, deep neural networks are known for their ability to analyze complex signals because they have successive layers with non-linear activation functions. The deepness of the network has been correlated with the capacity of the network to analyze very complex problems \citep{simonyan2014very, srivastava2015training}.} Second, hand-crafted methods are mostly inefficient in extracting discriminative gait features for a specific recognition problem. For example, in the work of \citet{ertuugrul2016detection}, LBP patterns were extracted and analyzed starting from VGRF signals. In contrast, the proposed model learns specific spatial kernels based on the data that allows the best distinction between the Parkinson and the control group. It is known that by increasing the model capacity, deep learning algorithms may make the classification vulnerable to overfitting. The algorithm may, in this case, memorize individual subject gait characteristics instead of memorizing discriminative patterns between the Parkison and control gaits. Nevertheless, our algorithm showed a high capacity to generalize to the validation set, which suggests that our 1D-Convnet has learned discriminative features between the parkinsonian and normal gait.

\subsection{Ablation study: VGRF signal selection}
In order to determine which VGRF time series has the most important impact on \textcolor{red}{Parkinson's disease detection}, we deleted each symmetric right and left VGRF pairs in turn (R1 \& L1 for example) and we trained the model with the remaining 16 VGRF signals using the same hyperparameters as before. 
\begin{table*}[hbt!] 
\centering 
\caption{Impact of input signals on Parkinson's detection \textcolor{red}{at segment level}. In each row, our DNN was retrained without (w/o) two symmetric inputs Right (R) and Left (L) VGRF signals. Results corresponding to the most significant features are in bold. \textcolor{brown}{Sp: Specificity, Se: Sensibility, Acc: Accuracy.}}
\begin{adjustbox}{width=\columnwidth,center}
\begin{tabular}{|l||c|c|c|}
\hline
 VGRF inputs &  Sp (\%)  & Se (\%) & Acc (\%) \\ \hline

w/o L1 \& R1  & 97.4& 93.7 & 94.8 \\ \hline
w/o L2 \& R2  &96.7 & 93.4 & 94.4 \\ \hline
w/o L3 \& R3 &  99.3 & 98.6 & 98.8 \\ \hline
w/o L4 \& R4 &  96.0 & 94.0 & 94.6 \\ \hline
w/o L5 \& R5 &\textbf{95.1} &95.0 & 95.1 \\ \hline
w/o L6 \& R6 &   97.3& 97.3& 94.8 \\ \hline
w/o L7 \& R7 &  96.2&  94.7& 95.2 \\ \hline
w/o L8 \& R8 & 96.5& 94.9 & 95.4 \\ \hline
w/o Total VGRF (R \& L) &  96.8 & \textbf{92.9} & \textbf{94.1}\\ \hline

\end{tabular}
\end{adjustbox}
\label{table:test_sel}
\end{table*}





The results of the VGRF signal selection are presented in table \ref{table:test_sel}. \textcolor{red} {Here, we present the results at the segment-level to facilitate the differentiation between the ablations. The deletion of features L2 \& R2, L4 \& R4 are the signal points that impacted the most the performance of our algorithm. Thus, we can hypothesize that the force measured at those points is different between the control and Parkinson group. However, the feature that seems the most important is the total VGRF. This feature is important in Parkinson's gait analysis, since most of the clinical features are generally extracted from the total VGRF. The deletion of features L3 \& R3 seems to not affect the performance of the algorithm. We can conclude then that this input signal was not relevant for the gait classification. }


\subsection{Parkinson severity prediction}

\begin{table*}[t]

\centering
\caption{Cross-validation results for Parkinson severity prediction (n: number of subjects) }
\begin{adjustbox}{width=\columnwidth,center}
\begin{tabular}{|l||c|c|c|c|}
\hline
 Class &  Precision (\%)  & Recall (\%) & F1 (\%) & n \\
\hline\hline
1  & 77.6 & 100.0 & 87.4 & 90  \\ \hline
2& 100 & 75.0 & 85.7 & 8  \\\hline
3 & 100.0 & 76.3 & 86.6 &76 \\\hline
4 & 91.8 & 80.0  & 85.5  & 70 \\\hline
5 & 78.0 & 82.1 & 80.0 & 56 \\\hline
Weighted average & 87.3 & 85.3 & 85.3 & 300\\\hline
\end{tabular}
\end{adjustbox}
\label{table:level}
\end{table*}

\begin{table*}[t]
\caption{Confusion matrix for Parkinson's severity prediction. All values are normalized with number of subjects in each category. }
\centering
\begin{tabular}{cc|ccccc|}
\multicolumn{1}{c}{} &\multicolumn{1}{c}{} &\multicolumn{5}{c}{Predicted  } \\ 
 &  & 1 & 2 & 3 & 4 & 5 \\ 
\hline
\multirow{5}{*}{\rotatebox{90}{Ground truth (\%)}} & 1 & 100 & 0.0 & 0.0 & 0.0 & 0.0   \\ 
 & 2 & 12.5 & 75.0 & 0.0 & 0.0 & 12.5   \\ 
 & 3 & 15.8 & 0.0 & 76.3 & 2.6  & 5.3  \\ 
 & 4 & 8.6 & 0.0 & 0.0 & 80.0 & 11.4 \\ 
 & 5 & 12.5 & 0.0 & 0.0 & 5.3 &  82.1  \\ 
\hline
\end{tabular}
\label{table:conf}
\end{table*}


\textcolor{red}{Table \ref{table:level} presents the results per class for the severity level prediction. The overall achieved accuracy is 85.3 \% (equivalent to the average recall). We can see that the F1 score is relatively similar between classes, so the algorithm is relatively consistent in his prediction between levels. Class 1 is the class where the algorithm was the most successful. This can be explained by the number of data available. In fact, this class contains all the control subjects. Nevertheless, the algorithm performed adequately and reaches a global accuracy of 85.3\%, even if the dataset is unbalanced.}

\textcolor{red}{From the confusion matrix presented in table \ref{table:conf}, we can also see that the vast majority of subjects for each class are correctly classified. However, the preponderant errors come from a confusion with class 1. Once again, this can be explained by the amount of data available in class 1 compared to other classes.}

\section{Conclusions}\label{sec:conc}

Parkinson diagnosis is still a very challenging problem in medicine. A Parkinson diagnosis is theoretically impossible to confirm, and doctors can detect the disorder by analyzing several symptoms through physical examination. \textcolor{blue}{Since gait perturbation is among the important motor symptoms}, we proposed \textcolor{red}{an algorithm to recognize the Parkinsonian gait and predict the severity of the disease based on gait data.} Our algorithm uses deep learning techniques, which avoids the drawbacks of hand-crafted feature extraction. The proposed DNN reached an accuracy of 98.7\% in Parkinson's gait recognition. \textcolor{red}{To the best of our knowledge, this represents the state-of-the-art performance. Furthermore, the proposed algorithm is the first to predict the UPDRS severity of a subject, with an accuracy of 85.3 \%}. This system can serve as a practical tool to screen a population in order to detect potential Parkinson patients in a clinical context. \textcolor{blue}{Moreover, we are witnessing an impressive democratization of biomedical sensors, that are becoming more and more present in our life. In the long-term, we believe that the proposed algorithm would be useful for the elderly, by monitoring and analyzing gait features during daily life activities. Such AI tools coupled with increasingly powerful biometric sensors would allow detecting gait abnormalities at the first stages of Parkinson’s disease.}

\textcolor{blue}{For future work, it would be interesting to get inside the DNN layers and analyze what they have learned. Such a study would allow a deeper understanding of the parkinsonian gait and its characteristics.}

\section*{Acknowledgement}

This research is partly funded by Fonds de Recherche du Quebec-Nature et Technologies (FRQ-NT) with grant No. 2016-PR-189250, and an institutional research fund from T\'{E}LUQ University. 



\bibliographystyle{model2-names}

\bibliography{refs.bib}





\end{document}